\def\year{2020}\relax
\documentclass[letterpaper]{article} 
\usepackage{aaai20}  
\usepackage{times}  
\usepackage{helvet} 
\usepackage{courier}  
\usepackage[hyphens]{url}  
\usepackage{graphicx} 
\usepackage{graphicx}  
\usepackage{algorithm}      
\usepackage[noend]{algpseudocode} 
\algnewcommand{\LeftComment}[1]{\Statex \(\triangleright\) #1}
\usepackage{subfigure}
\usepackage{amsmath}
\usepackage{amsfonts}
\usepackage{booktabs}
\usepackage{nicefrac}       
\usepackage{microtype}      

\title{A Forest from the Trees: Generation through Neighborhoods}
\author{Yang Li, Tianxiang Gao, Junier B. Oliva\\
Department of Computer Science, UNC Chapel Hill\\
\{yangli95, tianxiang, joliva\}@cs.unc.edu}

\begin{document}

\maketitle

\begin{abstract}
In this work, we propose to learn a generative model using both learned features (through a latent space) and memories (through neighbors). Although human learning makes seamless use of both learned perceptual features and instance recall, current generative learning paradigms only make use of one of these two components. Take, for instance, flow models, which learn a latent space that follows a simple distribution. Conversely, kernel density techniques use instances to shift a simple distribution into an aggregate mixture model. Here we propose multiple methods to enhance the latent space of a flow model with neighborhood information. Not only does our proposed framework represent a more human-like approach by leveraging both learned features and memories, but it may also be viewed as a step forward in non-parametric methods. In addition, our proposed framework allows the user to easily control the properties of generated samples by targeting samples based on neighbors. The efficacy of our model is shown empirically with standard image datasets. We observe compelling results and a significant improvement over baselines. Combined further with a contrastive training mechanism, our proposed methods can effectively perform non-parametric novelty detection.
\end{abstract}

\section{Introduction}

\begin{figure*}[htb]
\centering
\subfigure[Low Dimensional Neighborhoods]{
\includegraphics[width=0.45\linewidth]{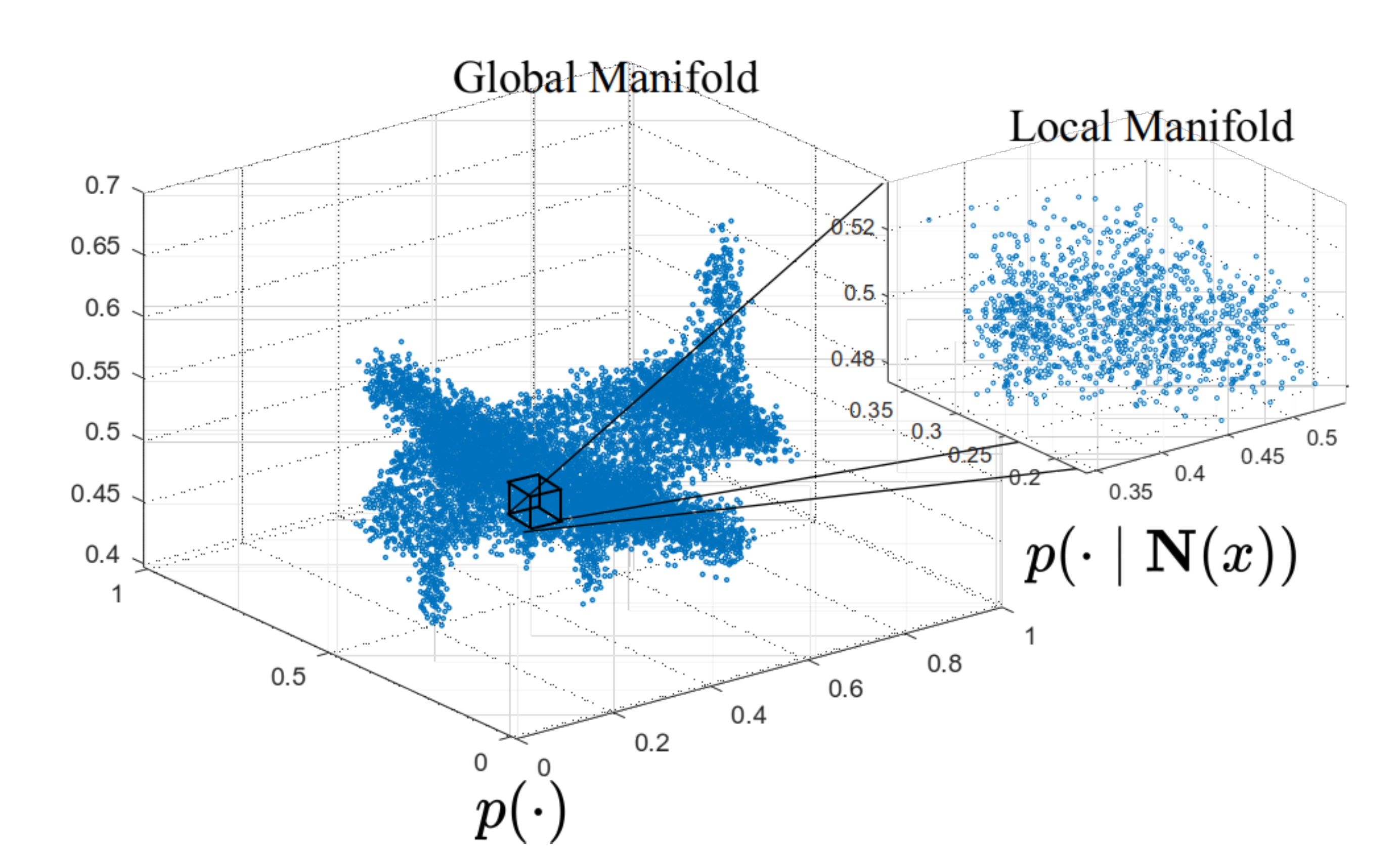}
\label{fig:kde}
}
\subfigure[High Dimensional Neighborhoods]{
\includegraphics[width=0.4\linewidth]{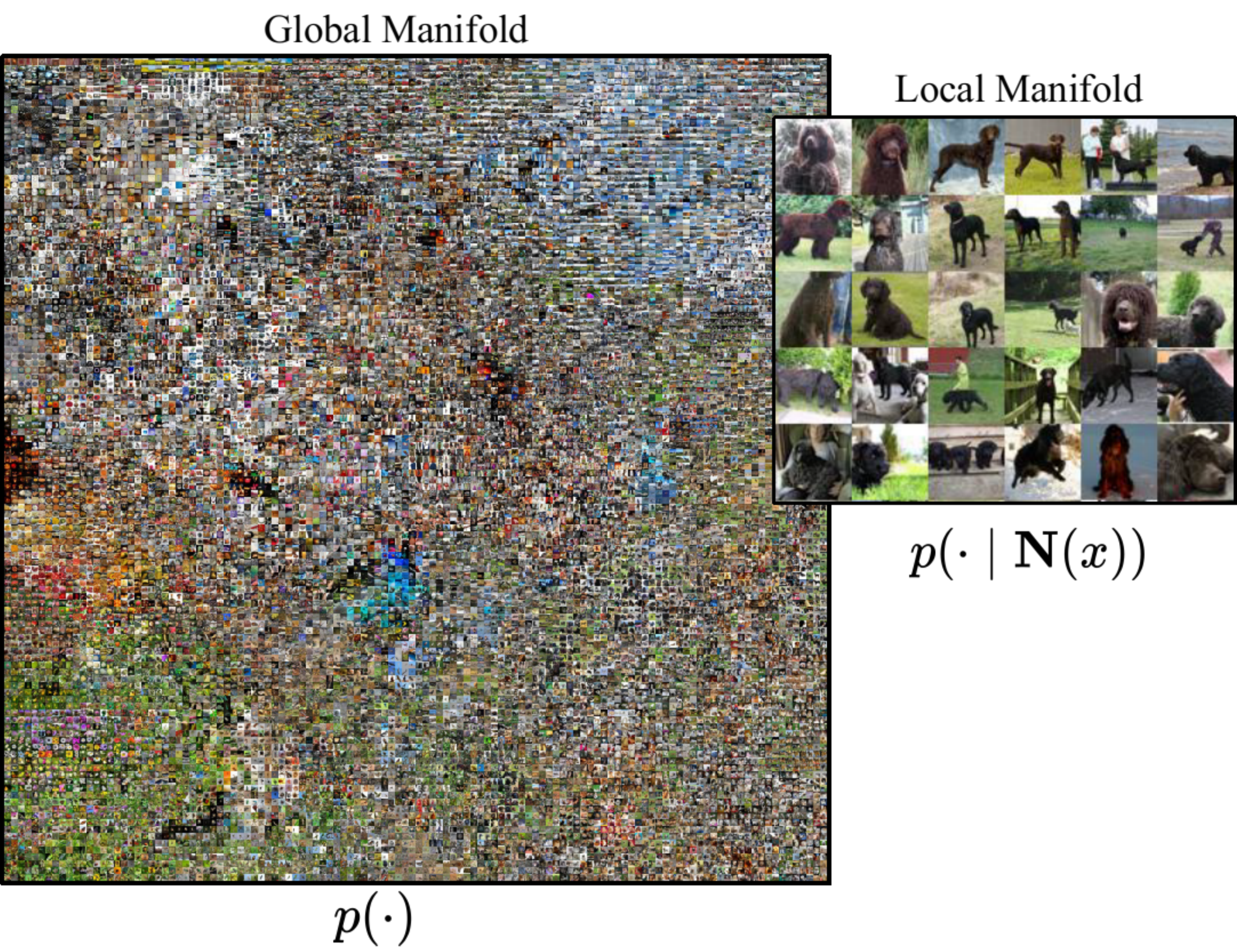}
\label{fig:img_neighs}
} 
\caption{
Neighborhood based models, which focus on modeling a simpler local manifold rather than the more complicated global support.
(a) In low dimensions, one may model tight local manifolds using simple distributions (e.g. Gaussians).
Thus, in low dimensions it is relatively simple to model a rich class of densities as the mixture of shifted simple distributions (i.e. kernels).
(b) In higher dimensions, such as the space of natural images, the number of simple kernels to adequately model the space grows exponentially, and is thus intractable.
Tractable local manifolds may no longer be Gaussian, however they still exhibit a structure that can be exploited.
In this work, we propose to model local manifolds using flexible density estimators and careful conditioning.
}
\label{fig:localmanifolds}
\end{figure*}

The typical training paradigm for many generative models makes a one time use of the training data.
For instance, VAEs \cite{kingma2013auto} and flow models \cite{dinh2014nice,dinh2016density,kingma2018glow} use the training data to fit a latent space. 
After the features (the latent space) are learned, the training data is discarded, as the learned network is entirely responsible for the generative process.
This paradigm has proven effective, however it leaves the entire burden of modeling a complicated support (such as the space of images) completely on the learned latent space and network capacity.
Moreover, such an approach is in stark contrast to human learning, which not only uses data to learn perceptual features \cite{kuhl2003foreign}, but will reuse data as memories with instance recall \cite{Carrier1992}.
Furthermore, since samples are generated from a random latent code, it is difficult to control the properties of generated samples. 
In this work, we more closely follow nature's paradigm for generative modeling, making use of our data not only to learn features but to pull instances as neighbors.

Neighborhood based methods abound much of the history of machine learning and statistics. 
For instance, the well-known $k$-nearest neighbor estimator often achieves surprisingly good results \cite{altman1992introduction}.
Moreover, by equipping models with neighborhood information, one reduces the problem to a local-manifold, a much simpler domain \cite{roweis2000nonlinear,seung2000manifold}.
In this work, we focus our attention on density estimation, which features a staple neighbor based estimator: the kernel density estimator (KDE).
The KDE models a local manifold as a Gaussian, an untenable restriction when dealing with complicated data such as images, see Fig.~\ref{fig:localmanifolds}.
Instead, we model local manifolds using more flexible densities, simultaneously better representing the local space of a neighborhood whilst reducing the burden of the density estimator.
In doing so, this work brings forth methodology that is in the same vein as non-parametric density estimation. 
We begin with a simple observation about the kernel density estimator.
Recall that KDE models a dataset $\{x_i \in \mathbb{R}^d\}_{i=1}^N$ as a mixture of ``kernel'' distributions $K$ centered around each data-point:
\begin{equation}
p(x) = \frac{1}{N} \sum_{i=1}^N K(x_i, x).
\end{equation}
In practice, the kernel is often chosen to be a Gaussian distribution centered at the data-points with a diagonal covariance matrix: 
\begin{equation}
p(x) = \frac{1}{N} \sum_{i=1}^N \mathcal{N}(x \mid x_i, \sigma^2 \textbf{I}), \label{eq:kde}
\end{equation}
where $\mathcal{N}(x \mid x_i, \sigma^2 \textbf{I})$ is the pdf of a Gaussian with mean $x_i$ and covariance matrix $\sigma^2 \textbf{I}$.
That is, the density is modeled as a mixture of training data-centered distributions.
The sampling procedure is simple, as one selects a training point $x_i$ from the training data at random and adds Gaussian noise to it.
More generally, one sees that KDE models the density as:
\begin{equation}
p(x) = \frac{1}{N} \sum_{i=1}^N p(x \mid x_i),
\end{equation}
for a very restricted class of conditionals $p(\cdot | x_i)$, namely shifted Gaussians.
We propose a data-driven generalization of KDE, another step forward in non-parametric methods.
We use neighborhoods to allow the estimator to model local manifolds, a simpler space. 
Specifically, we model the density as:
\begin{equation}
p(x) = \frac{1}{N} \sum_{i=1}^N p_\theta(x \mid \textbf{N}(x_i)), \label{eq:neigh_dens}
\end{equation}
where $p_\theta(\cdot \mid \textbf{N}(x_i))$ is \emph{a more flexible density estimator conditioned on neighborhood information} (Fig.~\ref{fig:img_neighs}) and $\textbf{N}(x_i)$ is the neighborhood around point $x_i$.

Our approach \eqref{eq:neigh_dens} is capable of inferring relevant features and variances of the neighborhood, lessening the burden on $p_\theta$.
Note further that KDE \eqref{eq:kde} blurs the line between density estimation and data augmentation; 
KDE samples by ``augmenting'' a data point with Gaussian noise.
The proposed approach expands on this by learning how to ``augment'' or sample given a neighbor or neighborhood.

The sampling procedure for our proposed model would remain simple: one would choose a training point $x_i$ uniformly at random from the training data, then sample according to the estimator conditioned on the neighborhood: $\mbox{$x \sim p_\theta(\cdot \mid \textbf{N}(x_i))$}$. 
\emph{Note that the entire sampling procedure is not a conditional model since we marginalize out the effect of the neighborhood by picking them from training data at random \eqref{eq:neigh_dens}.}

In addition to the simple global unconditioned sampling procedure, our paradigm also offers an unprecedented targeted sampling interface. Suppose that a user wanted to produce novel images of a tiger in a grassy outdoor setting, standing, and facing the camera. In order to provide a user with such fine-grained control in standard conditional models, one would have to train a multi-labeled model based on concepts such as ``contains tiger'', ``orient forward'', ``outdoor'', ``grassy'', etc. However, such a labeled dataset may be very expensive to collect. Furthermore, it may be hard to obtain a compact representation for certain concepts \cite{higgins2016beta}. Lastly, a multi-labeled conditional model would be unable to extrapolate to unforeseen concepts at test times. These shortcomings make a traditional approach unfit for a targeted sampling task. Instead, our neighbor based approach is very adept at producing targeted novel samples. In the aforementioned scenario, the user would be able to condition the neighborhood model using tiger images similar to his/her target. In effect, our model provides an indispensable ``more like this'' interface to generative models.

Our proposed approach builds on the standard non-parametric methodology in three key ways.
\emph{First}, we propose to use more robust density kernels than the standard approaches; while kernels like the Gaussian distribution lead to general distributions asymptotically, their simple unimodal nature often falls short for higher dimensions in finite samples \cite{Wasserman:2010:NS:1951569}.
\emph{Second}, we propose to condition on training data in a more robust fashion than by simply shifting a base distribution as is standard; this allows one to capture a richer set of correlations of data in a local neighborhood.
\emph{Third}, we propose to extract the set-level information from the neighborhood $\textbf{N}(x_i)$ in a data-driven fashion; conditioning on multiple instances in a neighborhood allows our model to learn the variances present in the local manifold (in the background of images, for instance) without the network needing to memorize them.

In all, our proposed paradigm makes use of both memories (via instance recall of neighborhoods) and learned features (via a latent space). As such, our approach is richer than existing methods, more closely resembles human learning, and gives us fine-grained control over sample properties. We show the efficacy of our models on standard benchmark image datasets with experiments below.

\section{Methods}\label{methods}

\begin{figure*}[htb]
\centering
\subfigure[RNVP]{
\label{fig:model.rnvp}
\includegraphics[height=0.35\textwidth]{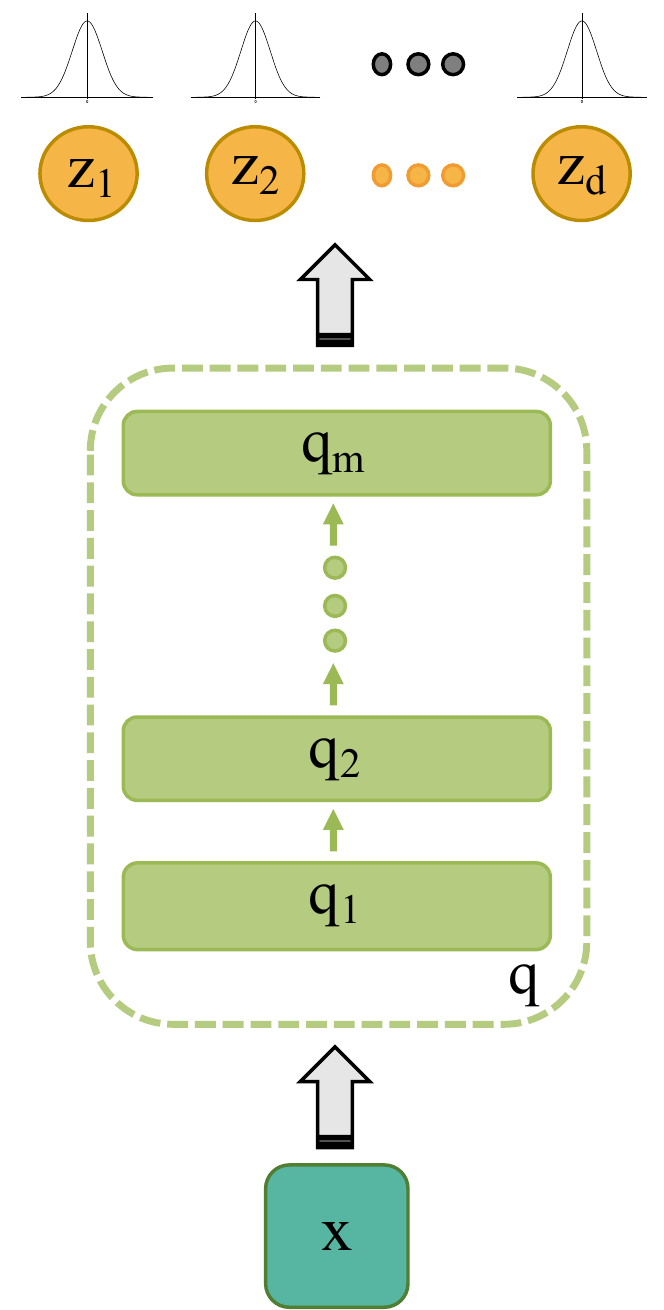}}
\qquad~~
\subfigure[NCL]{
\label{fig:model.nei}
\includegraphics[height=0.42\textwidth]{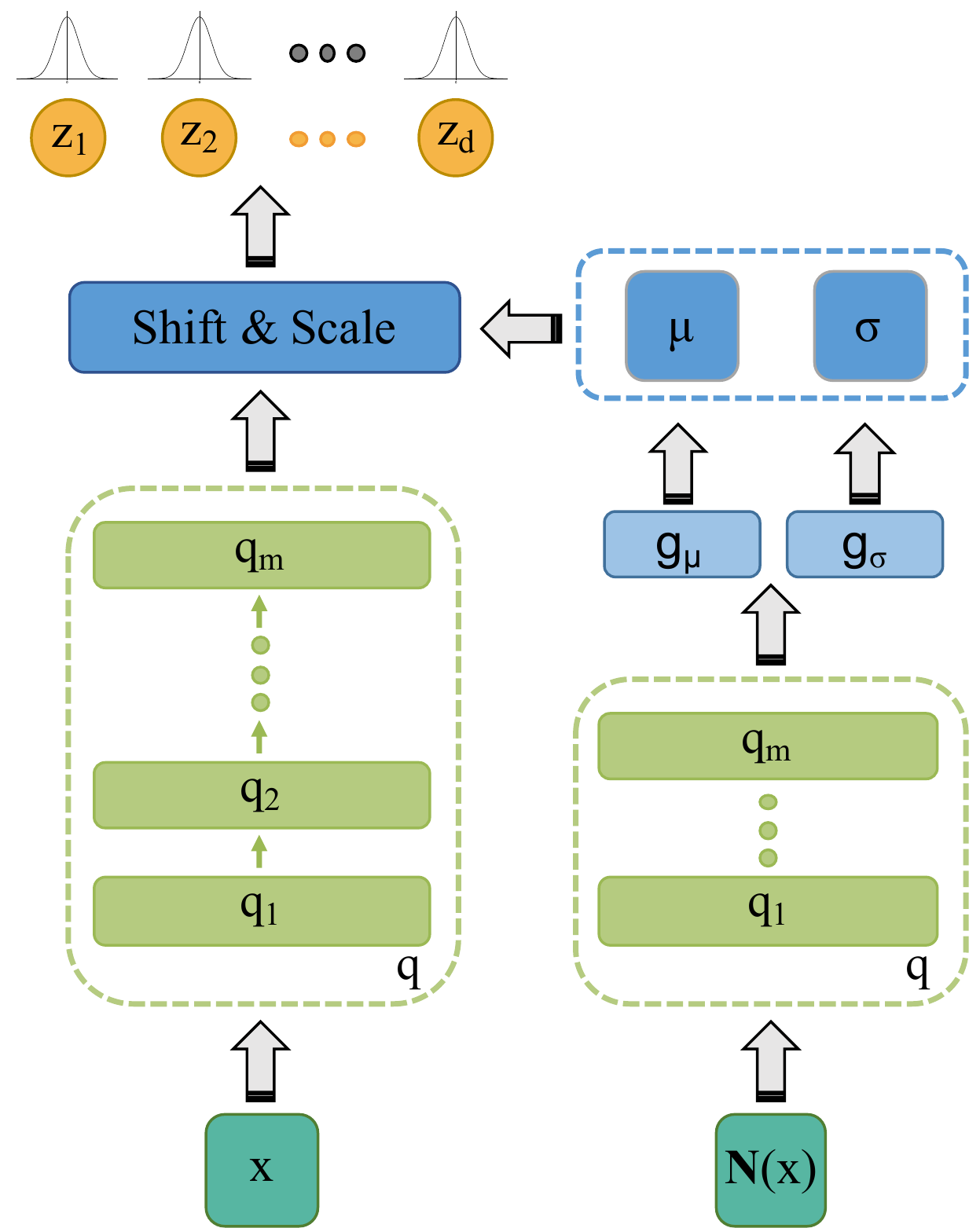}}
\qquad~~
\subfigure[NCT]{
\label{fig:model.nt}
\includegraphics[height=0.35\textwidth]{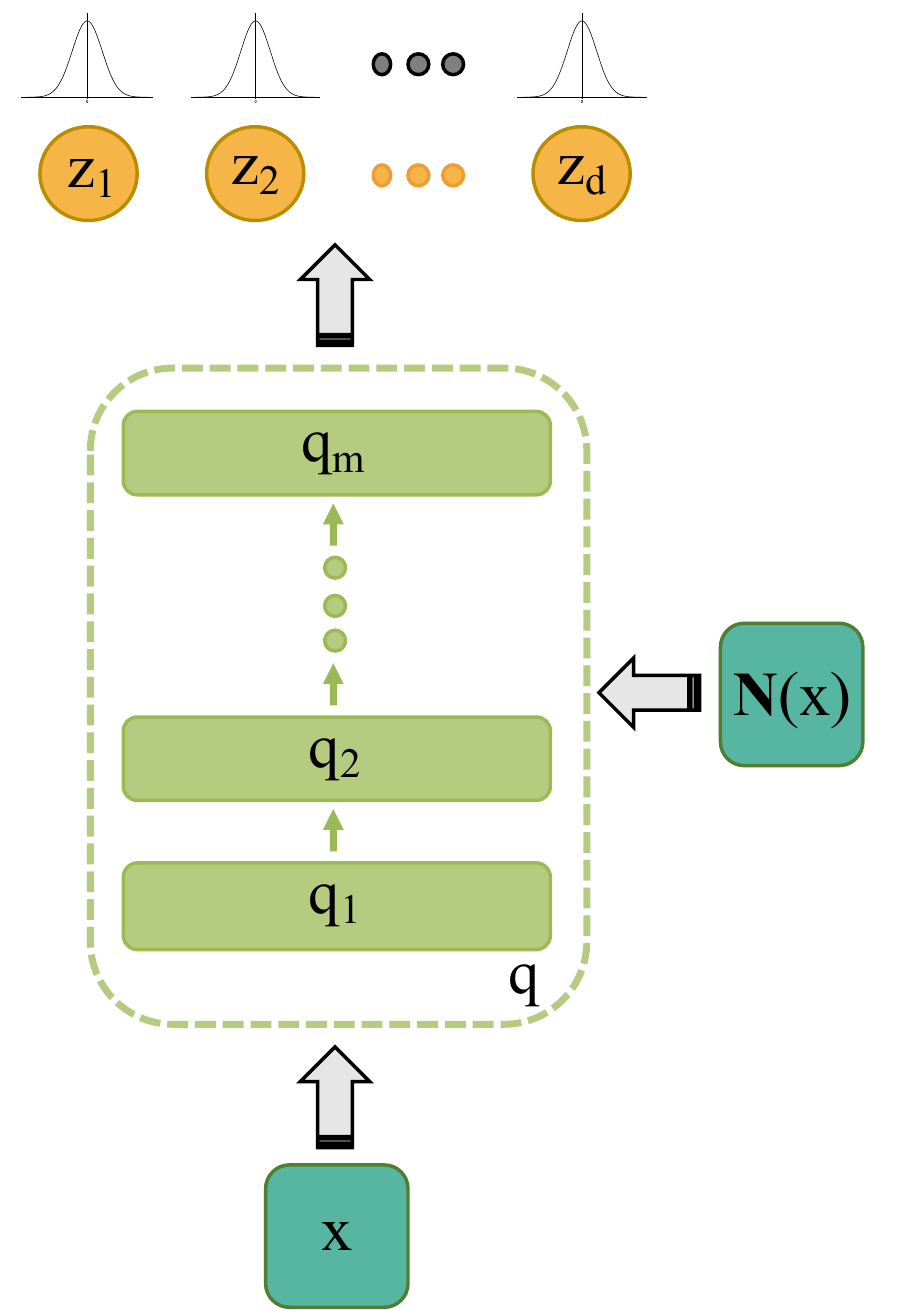}}
\caption{(a) Vanilla RealNVP model. The dashed box $q$ indicates a series of invertible transformations, which transform images $x \in \mathbb{R}^d$ (cyan blocks) into $d$ standard Gaussian covariates (yellow circles). (b) Conditioning latent distribution on neighbors. Note that the two dashed green boxes share the same group of parameters (same transformation for input images and corresponding neighborhoods). $g_{\mu}$ and $g_{\sigma}$ compute mean and variance based on the specified neighborhood respectively. (c) Using neighbors to condition the transformations. Some or all transformations in $q$ could depend on neighborhood $\textbf{N}(x)$.}
\label{fig:model}
\end{figure*}

While the local manifold of a neighborhood might be remarkably simpler than the global manifold, one is still unlikely to estimate the local manifold well with a simple base distribution such as a Gaussian.
Thus, we still need to capture the data in these local neighborhoods $\textbf{N}$ with a flexible model $p_\theta(\cdot \mid \textbf{N})$.
The choice of robust base conditional model is quite flexible, with many possible choices including autoregressive models \cite{oord2016pixel,van2016conditional}, flow models \cite{dinh2014nice,dinh2016density,kingma2018glow}, and variational autoencoders (VAEs) \cite{kingma2013auto}.
In fact, one may also deviate from a likelihood based framework in favor of a discriminative critic GAN approach \cite{goodfellow2014generative}.
As our focus is on the methodology of injecting neighborhood information into a generative approach, we explore this with a particular choice of generative model, flow density estimators. However, our approach easily extends to other generative models.
In addition, flow methods provide a few advantages including: a tractable (normalized) likelihood, robustness to mode collapse \cite{grover2018flow}, and a meaningful latent space. We expound on the base flow generative models below.

\subsection{Flow based Models}
The change of variable theorem, shown in \eqref{eqn:changevariable}, is the cornerstone of flow generative models \cite{dinh2014nice,dinh2016density,kingma2018glow}, where $q$ is an invertible transformation.
\begin{equation}
    p_X(x) = \left|\det\frac{dq}{dx} \right| p_Z(q(x)) 
    \label{eqn:changevariable}
\end{equation}
In order to efficiently compute the determinant, the transformation $q$ is often designed to have diagonal or triangular Jacobian. Since this type of transformation is restricted, the flow models often compose multiple transformations in a sequence to get a flexible transformation, i.e.  $q = q_m \circ q_{m-1} \circ \ldots \circ q_1$. Here, the covariates flow through a chain of transformations, substituting the last output variable as input for the next transformation.

One family of this type of transformation is the so-called \textit{coupling} layer used in NICE \cite{dinh2014nice}, RealNVP \cite{dinh2016density} and Glow \cite{kingma2018glow}. The input $x$ is divided into two parts $x^A$ and $x^B$, the first part is kept the same, and the second part is transformed based on the first part using an affine transformation,
\begin{align}
\begin{split}
    y^A &= x^A\\
    y^B &= x^B \odot s(x^A)+t(x^A),
\end{split}
\end{align}
where $\odot$ represents the element-wise (Hadamard) product.
Such coupling layers are easy to invert and the Jacobian determinant is easy to compute. Furthermore, since the Jacobian determinant does not involve computing the Jacobian of $s(\cdot)$ and $t(\cdot)$, they can potentially be any deterministic function, such as a neural network.

Since one single coupling layer can only affect part of the input covariates, earlier works swap $A$ and $B$ between two coupling layers to ensure every dimension can affect every other dimension. Recent work \cite{oliva2018transformation,kingma2018glow} proposes to learn a \textit{linear} layer to better capture the correlations along dimensions. In order to stabilize the training and ensure the whole process's invertibility, the RealNVP reformulates the batch normalization into an invertible transformation. They also propose a multi-scale architecture to capture spatial correlations in image data.

In the end, the flow models transform inputs to a new space, $z=q(x)$, where covariates can be modeled using a simple base distribution, see Fig.~\ref{fig:model.rnvp}. Typically, we take $p_Z(z)$ to be a Gaussian distribution. 
From another perspective, we use a deep neural network to construct a series of flexible and invertible operations to transform a simple distribution (e.g. a Gaussian) into a complicated one (the data distribution). 
Flow based generative models use the exact log-likelihood $\log~p(x)$ as the training criterion. 

\subsection{Neighbor Conditioned Flow Models}

We impose the neighborhood information in neighbor conditioned distributions \eqref{eq:neigh_dens} by conditioning a flow model.
Flow models can be divided into two parts, the transformation of variables and the latent distribution. Both of these parts can incorporate extraneous conditioning information. 
Concretely, the transformation procedure can use neighborhood information to decide how to transform the inputs, while the latent distribution can have mean and variance depending on conditioning variables. 
We explore two ways to inject neighborhood information into the flow model. 
First, we propose neighborhood conditioned likelihood (\textbf{NCL}), which specifies the distribution of the latent covariates of a target point, $z=q(x)$, given its neighborhood $\textbf{N}$.
Second, we propose neighborhood conditioned transformations (\textbf{NCT)}, which adjusts the latent space (transformation of variables) for a target point $x$ according to its neighborhood $\textbf{N}$, i.e., $z=q_\textbf{N}(x)$.

\paragraph{Neighborhood Conditioned Likelihood (\textbf{NCL})} We propose to directly estimate the density of the representation of a target point $x$ in the latent space, $q(x)$, given its neighborhood $\textbf{N}$.
To do so, two mappings, $g_\mu, $ and $g_\sigma$, are learned to estimate mean and variance parameters, $\vec{\mu}$ and $\vec{\sigma}$, respectively from $\textbf{N}$. 
We disentangle the transformation of variables from the relationship between the neighbors and the target by passing in transformed neighbors to $g_\mu, g_\sigma$, i.e., we pass neighbor images through the same flow transformations $q$. Then, we use $g_\mu$ and $g_\sigma$ to compute Gaussian means and variances respectively.
As in original flow models, we use a diagonal matrix for the variance:
\begin{equation}
    p_Z(z) = \mathcal{N}(z \mid g_{\mu}(q(\textbf{N})), diag(g_\sigma^2(q(\textbf{N})))),\label{eq:ncl}
\end{equation}
The model is illustrated in Fig.~\ref{fig:model.nei}. 
To sample, we choose a neighborhood $\textbf{N}$ uniformly at random from the training data. 
Then, we transform each neighbor image in $\textbf{N}$ into the latent space, and get the neighbor conditioned distribution based on latent codes of neighbors. 
We then sample latent codes, 
\mbox{$z \sim \mathcal{N}(\cdot \mid g_{\mu}(q(\textbf{N})), diag(g_\sigma^2(q(\textbf{N}))))$}, from this distribution and invert the flow transformation to get image samples, $x=q^{-1}(z)$. In another point of view, our \textbf{NCL} model resembles a KDE model in the latent space since we are sampling from a Gaussian conditioned on a neighborhood.
However, as we are operating in a latent space and are conditioning on the neighborhood according to the output of a general mapping, the \textbf{NCL} is a strictly more general model. 

\paragraph{Neighborhood Conditioned Transformations (\textbf{NCT})} 
As mentioned, another component of flow models that is amenable to conditioning information is the transformation of variables.
Although the \textbf{NCL} model allows the latent covariates $z$ to come from a distribution that depends on a neighborhood $\textbf{N}$, the construction of $z$ itself is uninformed by $\textbf{N}$.
We propose to inject neighborhood information into flow transformations using neighborhood conditioned transformations (\textbf{NCT}).
The \textbf{NCT} model provides guidance about how to effectively transform a local manifold through neighbors.
Here, we propose a neighborhood conditioned coupling transformation:
\begin{align}
\begin{split}
    y^A &= x^A\\
    y^B &= x^B \odot s(x^A,\textbf{N})+t(x^A,\textbf{N}).\label{eq:nct}
\end{split}
\end{align}
The shift and scale functions $s$ and $t$ are implemented via a concatenation operation on inputs. 
Replacing the coupling layer in a standard flow model with the proposed neighbor conditioned one gives rise to our \textbf{NCT} model.
The transformations \eqref{eq:nct} can model both intra-pixel and intra-neighbor dependencies.
Furthermore, by conditioning the flow transformations throughout each of the composing coupling transformation, we are able to inject our conditioning neighborhood information through the transformation of variables, rather than only at the end.
In cases of multi-scale architecture, like the one used in RealNVP, $x^A$ could have different spatial dimensions from $\textbf{N}$, thus we re-size the neighbors before concatenating them.

We note that the \textbf{NCL} model can be interpreted as a special case of \textbf{NCT}. One may view $g_\mu$ and $g_\sigma$ as specifying a shift and scale operation in the transformed space:
\begin{equation}
    z = \frac{q(x) - g_\mu(q(\textbf{N}))}{g_\sigma(q(\textbf{N}))},
\label{eq:ncl_as_nct}
\end{equation}
where $z$ now can be modeled as isotropic unit norm Gaussian. However, as we specify $g_\mu$ and $g_\sigma$ in terms of the transformation $q$, we get distinct models stemming from each approach. The details about our neighbor conditioned flow models are presented in Alg.~\ref{alg:eval} and Alg.~\ref{alg:sample} in Appendix.~\ref{sec:pcode}.

\subsection{Contrastive Training}
A straight-forward way to train our neighbor conditioned models is to directly optimize the neighbor conditioned likelihood. 
However, likelihood trained models are known to suffer from out-of-distribution (OoD) issues \cite{nalisnick2018deep,choi2018generative}, which means such models may assign a higher likelihood to data from a completely different distribution than data it is trained on. 
When deploying the models to perform novelty/anomaly detection, the insensitivity to OoD data can completely devastate the prediction. 
Inspired by the noise contrastive estimation (NCE) \cite{gutmann2010noise}, we propose a contrastive training mechanism to increase the sensitivity of our models to OoD data. 
A common and major challenge with NCE is the choice of the noise distribution and how to obtain “negative” samples. 
Fortunately, our neighborhood based paradigm provides a natural answer. 
For a given neighborhood \textbf{N}, $p(\cdot \mid \textbf{N})$ is to have support only in the local manifold around \textbf{N}. 
Hence, we can take $p(\cdot \mid \neg \textbf{N})$ for a different neighborhood to be a noise distribution w.r.t. $p(\cdot \mid \textbf{N})$ and take $x \in \neg \textbf{N}$ as negative samples.

That is, given $x_i$ and $x_j$ from two different neighborhoods $\textbf{N}(x_i)$ and $\textbf{N}(x_j)$ respectively, we optimize the following objective function:
\begin{equation}
\begin{aligned}
    \mathcal{L}_\theta = \mathop{\mathbb{E}}_{x \sim p(\cdot \mid \textbf{N}(x_i))} \mathop{\mathbb{E}}_{x^\prime \sim p(\cdot \mid \textbf{N}(x_j))} [ p_\theta(x \mid \textbf{N}(x_i)) \\
    + \max\left(0, m+p_\theta(x^\prime \mid \textbf{N}(x_i))-p_\theta(x \mid \textbf{N}(x_i))\right) ],
\end{aligned}
\end{equation}
where $m$ represents a margin between in-neighborhood likelihood and out-of-neighborhood likelihood. Although the original NCE objective uses an infinite margin, we find this finite margin objective stabilizes the training.

\section{Experiments}


Following the exact preprocessing procedure in RealNVP, we transform the pixels into logit space to alleviate the impact of boundary effects. 
In order to conduct a fair comparison with RealNVP, we use exactly the same network architecture and hyperparameters as those in \cite{dinh2016density}. 
We do not apply data augmentation for all experiments, but we use early stopping to prevent overfitting. 
We optimize the likelihood objective with contrastive training in novelty detection experiments.

In order to avoid cheating, we pull neighbors only from training set. 
Please refer to Appendix.~\ref{sec:nearest_neigh} for more information about getting neighborhoods. 
To explore the sensitivity of our methods to the choice of neighborhoods, we also inspect the behavior of our models using far fewer neighborhoods, where we cluster the training data into several neighborhoods, see Appendix.~\ref{sec:cluster} for more details.

In addition to vanilla RealNVP, we also compare to a simple class label/attributes conditioned RealNVP model. 
We model the latent space as a class/attributes conditioned Gaussian distribution. 
We use one fully connected layer to derive the Gaussian mean and variance respectively. 
In contrast to the Glow model \cite{kingma2018glow}, we do not share the class conditioned Gaussian over spatial dimensions, which we observed made our class conditioned model a much stronger baseline. 

When training based on the likelihood for a specific data point $x_i$, we optimize log likelihood values conditioned on the respective neighborhood, i.e., $\log p_\theta(x_i \mid \textbf{N}(x_i)).$
While the likelihood of the generative process is \eqref{eq:neigh_dens}, optimizing our neighborhood criteria encourages likelihoods to be concentrated around the respective conditioning neighborhood.
FID scores \cite{heusel2017gans} and precision-recall (PRD) \cite{sajjadi2018assessing} are used to quantitatively compare the sample quality. 
As in \cite{sajjadi2018assessing}, we report $(F_8, F_{\frac{1}{8}})$ pair to represent the recall and precision respectively. 
Models are compared using 50K samples.

\begin{table}[t]
\caption{FID scores. GAN results are from \cite{lucic2018gans,heusel2017gans}. Best scores are marked in \textbf{bold}.}
\label{tab:fid}
\centering
\small
\begin{tabular}{l|c|cc|ccr}
\toprule
        Dataset      &  GAN  & RNVP  & CC-RNVP  & NCL & NCT \\
\midrule
    MNIST     &  \emph{6.7}  & 10.4  & 8.3      & 11.2     & \textbf{8.2}      \\
    SVHN      &  \emph{12.5} & 103.8 & 55.2     & \textbf{50.0}     & 61.8     \\
    CIFAR-10  &  \emph{55.2} & 99.9  & 93.4    & 95.8     & \textbf{81.1}     \\
    CelebA    &  \emph{30.0} & 39.2  & 32.5 & 33.0     & \textbf{30.9}     \\
\bottomrule
\end{tabular}
\end{table}
\normalsize

\begin{table*}[t]
\caption{PRD scores. As in \cite{sajjadi2018assessing}, $(F_8, F_{\frac{1}{8}})$ pairs are reported here. GAN results are from \cite{sajjadi2018assessing} (marker B, i.e., the one with best precision and recall). Best scores for likelihood methods are in \textbf{bold}.}
\label{tab:prd}
\centering
\begin{tabular}{l|c|cc|ccr}
\toprule
     Dataset        & GAN  &  RNVP  &  CC-RNVP  &  NCL  &   NCT \\
\midrule
     MNIST          & \emph{(0.99, 0.99)}  & (0.982, 0.982) & (0.987, 0.973) & (0.981, 0.970) & (\textbf{0.990}, \textbf{0.982})\\
     SVHN           & \emph{(---,---)} & (0.740, 0.731) & (0.785, 0.920) & (0.808, \textbf{0.928}) & (\textbf{0.841}, 0.841)\\
     CIFAR-10       & \emph{(0.77, 0.84)}  & (0.322, 0.631) & (0.360, 0.599) & (0.451, 0.647) & (\textbf{0.568}, \textbf{0.691})\\
     CelebA         & \emph{(0.87, 0.85)}  & (0.675, 0.766) & (0.835, 0.786) & (\textbf{0.853}, 0.806) & (0.849, \textbf{0.821})\\
\bottomrule
\end{tabular}
\end{table*}

\subsection{Results and Analysis}

\begin{figure*}[t]
\centering
\subfigure{
\label{result:mnist}
\includegraphics[width=0.9\linewidth]{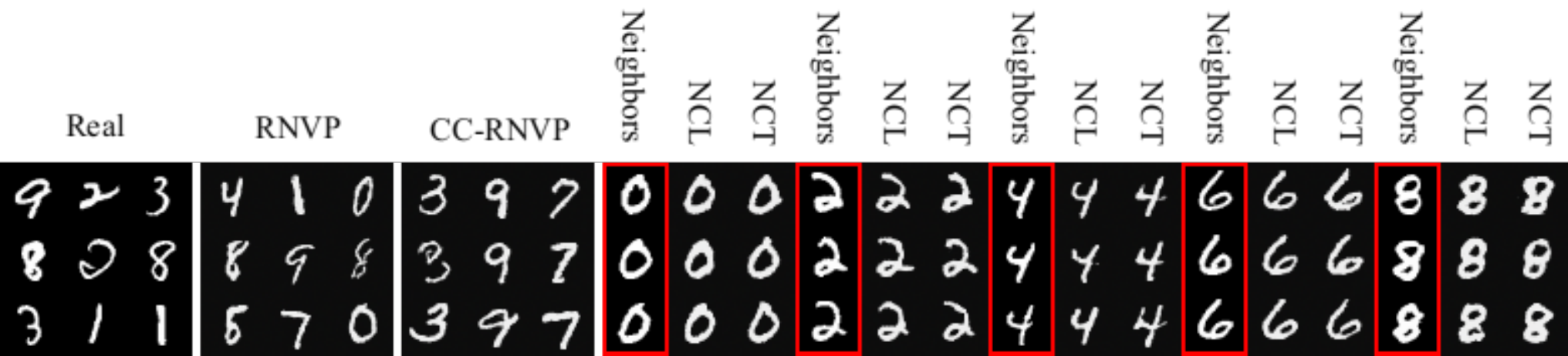}}
\subfigure{
\label{result:svhn}
\includegraphics[width=0.9\linewidth]{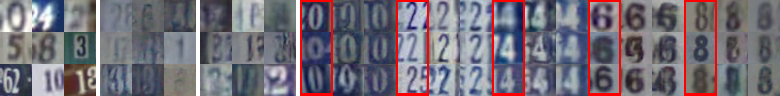}}
\subfigure{
\label{result:cifar}
\includegraphics[width=0.9\linewidth]{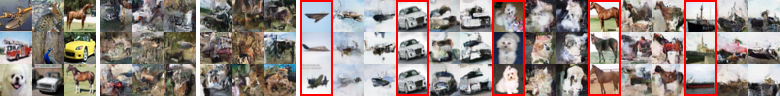}}
\subfigure{
\label{result:celeba}
\includegraphics[width=0.9\linewidth]{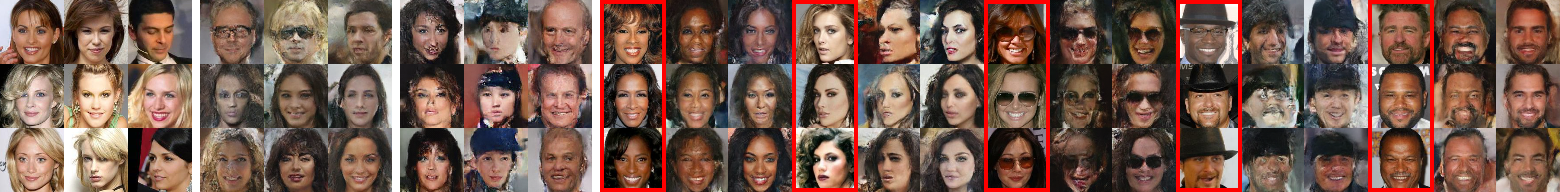}}
\caption{Samples generated form MNIST, SVHN, CIFAR-10 and CelebA. Columns in order are: real images, vanilla RealNVP (RNVP) samples, class conditioned RealNVP (CC-RNVP, each column represents one class) samples, and our neighbor conditioned models' samples. The red boxes indicate representative neighbors. The columns following each neighborhood column are samples from that neighborhood. The first column is from NCL, and the second one is from NCT.}
\label{fig:result}
\end{figure*}

We show samples generated for all four datasets in Fig.~\ref{fig:result}. We also show actual images in the first column for comparison. 

On all datasets we see that our models produce coherent samples with respect to the conditioned neighbors.
For example, on MNIST our neighbor based models allow fine-grained control over properties of samples, such as writing style and the brush width.
In SVHN, the vanilla RealNVP yields lesser quality samples, and the class conditioned RealNVP fails to generate samples concentrated around a single class. 
In contrast, our neighborhood based samples capture the attributes specified by their neighborhood very well. For instance, the background, the font and the character width closely match with the neighborhood. 
On CIFAR-10, our model generates similar images to the specified neighbors, while the vanilla RealNVP and class conditioned RealNVP fail to generate any meaningful samples. We still observe decent coherence with neighbors using our models even though neighbors considered in this dataset are more diverse, especially in the background (due to the small dataset size of CIFAR-10).
We note that the CelebA dataset comes with comprehensive attribute representations.
Thus samples from attributes conditioned model tend to align with the specified attributes. 
However, our targeted sampling gives fine-grained control over the sample properties without the expensive (to collect) attribute labeling data. We can see that our samples capture the high-level attributes of neighbors, such as the orientation, the ornament, the hair style and the skin color. 

Table~\ref{tab:fid} and Table~\ref{tab:prd} quantitatively compare the sample quality using FID and PRD respectively. We also list the FID scores of GANs for an additional comparison.
We see consistent improvement over the baseline RealNVP on both recall and precision. That is, our methods utilize neighborhood as memory recall to cover diverse data space and model local manifold to better capture the data manifold.
Our neighbor based targeted sampling serves to bridge the gap between non-adversarial and adversarial generative models.

\begin{table}[h!]
\caption{Bits per dimension results for MNIST, SVHN, CIFAR-10 and CelebA. Neighborhood conditioned likelihoods $p(\cdot \mid \textbf{N})$ are reported for our NCL and NCT models, while likelihoods and class conditioned likelihoods are reported for vanilla RealNVP (RNVP) and class conditioned RealNVP (CC-RNVP) respectively.}
\label{tab:results}
\centering
\begin{tabular}{l|cc|cccr}
\toprule
 Dataset  &  RNVP & CC-RNVP & NCL & NCT \\
\midrule
MNIST & 0.449 & 0.415 & 0.402 & 0.424\\
SVHN & 2.270 & 2.209 & 2.204 & 2.285\\
CIFAR-10 & 3.547 & 3.544 & 3.552 & 3.543\\
CelebA & 3.018 & 3.150 & 2.931 & 2.934\\
\bottomrule
\end{tabular}
\end{table}

Relative likelihood values can be found in Table.~\ref{tab:results}. 
We report bits per dimension (bpd) \cite{papamakarios2017masked} for the models considered. Note that we are not trying to compare conditional likelihoods with marginal likelihoods. We report likelihoods here only as a justification that our model is not trying to remember the training set. 

Furthermore, it is interesting to note that an approximation to the marginal likelihood for our model is typically close in value to conditional neighborhood likelihoods, see Appendix~\ref{sec:likel} for details.
We also explore reducing the computational cost of marginalization by using fewer neighborhoods.
To do so, we cluster the training data into several larger neighborhoods. 
It is worth noting that our models are robust to the choice of neighborhoods, and even with far fewer neighborhoods, we observe consistent samples with the corresponding neighborhood. Please refer to Appendix~\ref{sec:cluster} for more details.

\subsection{Novelty Detection}
Generative models have shown potential for novelty/anomaly detection, an important task in ML.
Although OoD issues can potentially limit the effectiveness of likelihood based methods, we find that our neighbor based models can significantly improve detection accuracy when combined with the proposed contrastive training strategy. 

Following the standard setup in \cite{perera2019ocgan}, we create one-class novelty detection tasks form MNIST dataset. We use the original training-testing split of MNIST dataset. Training is conducted using data of a single class from original training split. We test each model using the whole testing split. Test data from a different class the model is trained on are considered to be negative. To train our neighbor based models, we pull neighbors by Euclidean distance on PCA features, 5 neighbors are used here. Following previous works, we use Area Under the Curve (AUC) of Recevier Operating Characteristic (ROC) to measure the performance. We threshold the conditional likelihood $p(x \mid \textbf{N}(x))$ to get the ROC curve. In this experiment, we compare the vanilla RealNVP and our proposed \textbf{NCL} model with or without contrastive training. The margin in contrastive loss is set to 0.5 bits per dimension. We also list results of PixelCNN from \cite{perera2019ocgan} for an additional comparison. Results are shown in Table.~\ref{tab:novelty}. We can see that the proposed contrastive training mechanism contributes a huge improvement over NCL.

\begin{table*}[h]
\caption{Novelty detection results on MNIST dataset. Performance are measured by AUC-ROC. Best scores are in \textbf{bold}.}
\label{tab:novelty}
\centering
\begin{tabular}{lcccccccccccr}
\toprule
              & 0  & 1 & 2 & 3 & 4 & 5 & 6 & 7 & 8 & 9 & Mean\\
\midrule
    PixelCNN & 0.531 & 0.995 & 0.476 & 0.517 & 0.739 & 0.542 & 0.592 & 0.789 & 0.340 & 0.662 & 0.618 \\
     RNVP     & 0.417 & 0.996 & 0.528 & 0.547 & 0.726 & 0.598 & 0.665 & 0.814 & 0.420 & 0.703 & 0.641\\
     NCL(w/o) & 0.402 & \textbf{0.998} & 0.475 & 0.515 & 0.716 & 0.604 & 0.639 & 0.809 & 0.360 & 0.690 & 0.621\\
     NCL(w)   & \textbf{0.800} & 0.997 & \textbf{0.661} & \textbf{0.794} & \textbf{0.796} & \textbf{0.701} & \textbf{0.891} & \textbf{0.893} & \textbf{0.571} & \textbf{0.908} & \textbf{0.801}\\
\bottomrule
\end{tabular}
\end{table*}

\section{Related Works}

\textbf{Density Estimation and Generative Models:}
Nonparametric density estimation, such as kernel density estimation, often suffers from the curse of dimensionality and does not perform well on high dimensional data like images. Recently, deep neural networks have been employed to enable flexible density estimation. \cite{uria2013rnade,uria2016neural,germain2015made,gregor2013deep,oord2016pixel,van2016conditional} utilize neural networks to learn the conditionals factorized by the chain rule. \cite{dinh2014nice,dinh2016density,kingma2018glow,papamakarios2017masked,oliva2018transformation} construct a normalizing flow based on the change of variables theorem. \cite{kingma2013auto} proposes and optimizes a variational lower bound for the exact likelihood. GANs \cite{goodfellow2014generative} bypass explicit density estimation by adversarial training rooted in game theory. All of the aforementioned approaches try to model the whole data distribution in a single model. Our method, however, proposes to divide and conquer the density estimation using local neighborhoods. Our proposed model can potentially be integrated into all the generative models described above.

\textbf{Neighbors based Generative Models:}
We expound on some recent approaches that make use of neighbors for generative models below. 
For instance, \cite{bansal2017pixelnn} completes a low-resolution signal using compositions of nearest pixels in training images.
\cite{li2018implicit} attempts to model a fixed code space by matching initial noisy outputs to a nearest neighbor. Note that in contrast, in addition to providing a density, our method enables targeted sampling by specifying a neighborhood.

\textbf{Other Neighbor based Models:}
Outside of generative models, neighbor based methods have been well studied. For instance, \cite{weinberger2009distance,goldberger2005neighbourhood} apply nearest neighbors to learn a distance metric. \cite{boiman2008defense} extends classic KNN methods to a Bayes setting to perform accurate image classification. 
For a general discussion on nonparametric and neighborhood based methods please refer to \cite{Wasserman:2010:NS:1951569}.

\section{Conclusion}
In this work, we propose multiple ways of enhancing generative models with neighborhood information. 
Instead of modeling the whole manifold using a single model, we divide the support into smaller neighborhoods and model the simpler local manifolds.
Moreover, our approach jointly leverages the data both to learn a latent feature space and to use as neighbors to condition on a local manifold.
This reduces the burden on the network capacity as the model need not memorize the manifold properties it is conditioned on.
Furthermore, our approach more closely resembles human learning, which seamlessly leverages data to learn perceptual features and to recall instances as memories.

We extend the recently proposed RealNVP and propose two neighbor conditioned RealNVP architectures to model the local distributions. The neighborhood conditioned likelihood (\textbf{NCL}) models the latent distribution as a Gaussian conditioned on features of the neighborhood. In contrast, neighborhood conditioned transformations (\textbf{NCT}) adjust the latent space based on a neighborhood.

Empirical results show that the proposed neighborhood conditioned models improve the sample quality both quantitatively and qualitatively. Our training procedure yields models with a strong coherence between samples and neighborhoods, allowing for use in targeted sampling tasks. Furthermore, these models have the potential in neighborhood interpolation and style transfer as shown in our interpolation experiments Fig.~\ref{fig:interp}. Our models have the ability to generate realistic images even using far fewer neighborhoods. Combined with the contrastive training mechanism, our neighbor conditioned model significantly improves the novelty detection performance over the vanilla RealNVP.


\bibliographystyle{aaai}
\bibliography{AAAI-LiY.6146}

\newpage
\clearpage
\appendix
\renewcommand\thefigure{\thesection.\arabic{figure}}
\setcounter{figure}{0} 
\renewcommand\thetable{\thesection.\arabic{table}}
\setcounter{table}{0}

\section{Algorithms} \label{sec:pcode}

\begin{algorithm}[htb]
    \caption{Evaluate the neighbor conditioned flow model.}
    \label{alg:eval}
\begin{algorithmic}[1]
    \item[\textbf{Input:}] data $x_i$, neighbors $\textbf{N}(x_i)$, a sequence of invertible transformations $q$
    
    \item[\textbf{Output:}] log likelihood value $ \log~p(x_i \mid \textbf{N}(x_i))$
    
    \State Compute latent code $z_i$ for $x_i$ and the corresponding log Jacobian determinant (logdet) of the transformation:
    $[z_i, \log\det] = q(x_i \mid \textbf{N}(x_i)).$
    
        \LeftComment{For \textbf{NCL} type model, $q$ represents a sequence of transformations plus a neighbor conditioned shift and scale operation \eqref{eq:ncl_as_nct}.}
    
        \LeftComment{For \textbf{NCT} type model, $q$ represents a sequence of neighbor conditioned transformations \eqref{eq:nct}.}
    
    \State Evaluate the latent code in a isotropic unit norm Gaussian and get the log likelihood:
    $\log~p(x_i \mid \textbf{N}(x_i)) = \log~\mathcal{N}(z_i \mid \textbf{0}, \textbf{I}) + \log\det$
\end{algorithmic}
\end{algorithm}
    
\begin{algorithm}[htb]
    \caption{Sample from the neighbor conditioned flow model.}
    \label{alg:sample}
\begin{algorithmic}[1]
    \item[\textbf{Input:}] a sequence of invertible transformations $q$
    
    \item[\textbf{Output:}] a random sample $x_{sample}$
    
    \State Randomly sample a training instance $x_i$ and get a neighborhood $\textbf{N}(x_i)$ around it.
    
    \State Sample from a isotropic unit norm Gaussian: $z_{sample} = \mathcal{N}(z \mid \textbf{0}, \textbf{I})$
    
    \State Invert the transformation: $x_{sample} = q^{-1}(z_{sample} \mid \textbf{N}(x_i))$
\end{algorithmic}
\end{algorithm}

\section{Additional Experimental Details} \label{sec:exp_details}

\subsection{Implementation}

For class/attributes conditioned model, unlike Glow, we do not use the classification loss.
$g_\mu$ and $g_\sigma$ in \textbf{NCL} model are specified as two additional branches of six-layer convolution, and they share the first three layers. Features form neighbors are concatenated along the channel dimension after the first three convolutional layers.
Our code is available at \url{https://drive.google.com/file/d/1Az-jYn_caEKzgmUG9P0BKwlDppndBsOj/view?usp=sharing}.

\subsection{Nearest Neighborhood} \label{sec:nearest_neigh}

We compare our models with RealNVP model on four standard public datasets: MNIST, Street View House Number (SVHN), CIFAR-10 and CelabFaces Attributes (CelebA). For CelebA, same as \cite{dinh2016density}, we take a central crop of $148 \times 148$ then resize it to $64 \times 64$.
To get the neighbors for each dataset, we use PCA to reduce the dimension. We let the principal components explain 99\% of the variance on the training set. Then we search k nearest neighbors based on the PCA coefficients. For CelebA, the neighbors are pulled based on a pretrained VAE model. Here we use $\beta$-VAE \cite{higgins2016beta} and set $\beta$ equal to 1.5. We observe greater consistency between neighbors in this setting. 
For datasets with class labels, we restrict the neighbors in the same class to match human perception. We use 10 nearest neighbors for CIFAR-10 dataset and 5 for other datasets.
In all experiments, we pre-compute the neighbors and keep them fixed throughout the whole process. Though the neighbor searching strategy is very simple, it is capable of generating compelling samples.

\subsection{Likelihood Values} \label{sec:likel}

\begin{table}[h!]
    \centering
    \caption{Comparison between neighborhood conditioned likelihood and the full generative likelihood. Results are presented in bpd.}
    \label{tab:likeli}
    \begin{tabular}{lcccr}
    \toprule
          & $p_{\theta}(x \mid \textbf{N}(x))$ &  $\frac{1}{N}\sum_{i=1}^{N}p_\theta(x\mid \textbf{N}(x))$\\
    \midrule
    MNIST & 0.462 & 0.464\\
    Fashion & 1.082 & 1.080\\
    SVHN & 2.213 & 2.210\\
    CIFAR-10 & 3.030 & 3.027\\
    CelebA & 2.930 & 2.927\\
    \bottomrule
    \end{tabular}
\end{table}

As discussed above, the likelihood for the generative process is \eqref{eq:neigh_dens}. Here, we approximate the generative likelihoods by sampling additional 100 neighborhoods from the training set and compare to the conditional bpd in Table.~\ref{tab:likeli}. We average over 1000 samples from the test set to compute the conditional likelihood and the marginal likelihood. We notice that values are on par with each other. 
We hypothesize that the likelihood values tell a limited story due to ``out of distribution'' (OoD) issues with generative models. Recent works \cite{nalisnick2018deep,choi2018generative} have shown that density estimators like the RealNVP are unable to reflect whether an instance is a sample through likelihood values. For example, a RealNVP model trained on CIFAR-10 will return higher likelihood values for street view house number (SVHN) images than for CIFAR-10 natural images. As any instance outside of the local manifold is OoD for $p(\cdot \mid \textbf{N}(x_j))$, the systemic OoD discrepancy for these models represents a lot of ``wasted'' likelihood.

\subsection{Neighborhood Interpolation}

\begin{figure*}[!h]
    \centering
    \includegraphics[width=0.8\linewidth]{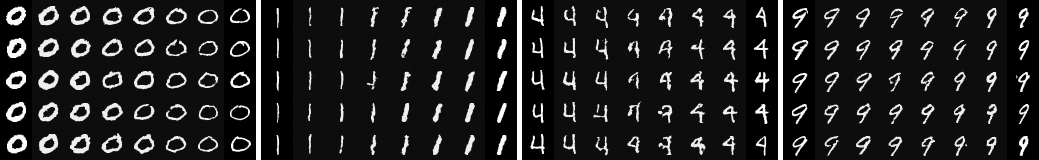}
    \caption{Interpolation between neighborhood. For each block, leftmost and rightmost columns are selected neighborhoods from training set. We gradually swap elements between them and sample from the ``interpolated'' neighborhood. Intermediate columns are samples.}
    \label{fig:interp}
\end{figure*}

In order to investigate how neighbors can affect the generation, we ``interpolate'' between neighborhoods. Specifically, we select two neighborhoods and gradually swap elements between them, reducing and increasing the number of elements by one for the neighborhood from the first and second set, respectively, at each step. The samples from the NCL-RNVP model are shown in Fig.~\ref{fig:interp}. We see a smooth change of attributes, such as the brush width, the orientation, and the writing style. 
Furthermore, as intermediate, interpolated, conditioning sets are unlike neighborhoods encountered at training time, this task showcases our models' ability to generalize to new neighborhoods.

\subsection{Neighborhood based on Clustering}\label{sec:cluster}

\begin{figure*}[!h]
\centering
\subfigure[MNIST]{
\includegraphics[width=0.45\textwidth]{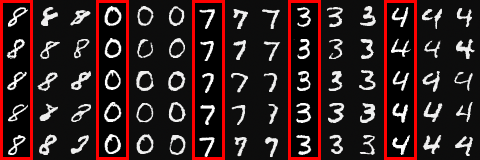}}
\subfigure[SVHN]{
\includegraphics[width=0.45\textwidth]{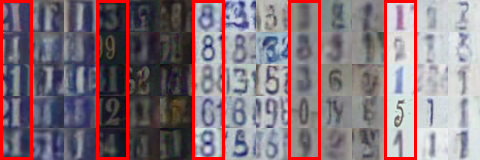}}
\subfigure[CIFAR-10]{
\includegraphics[width=0.45\textwidth]{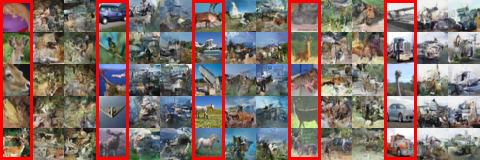}}
\subfigure[CelebA]{
\includegraphics[width=0.45\textwidth]{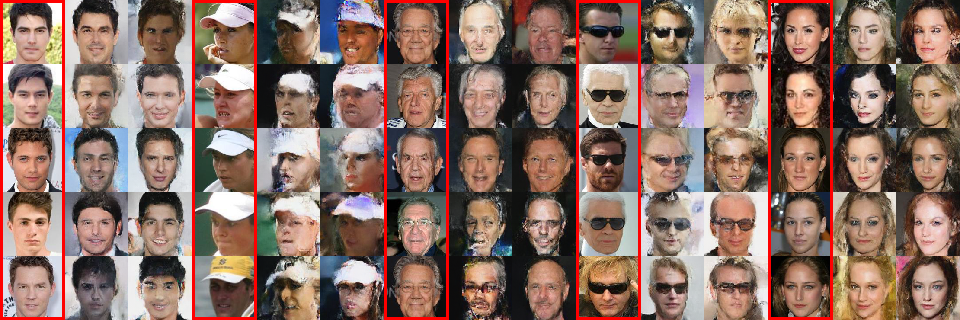}}
\caption{Samples generated from \textbf{NCT} model using clustering based neighborhood. The red boxes indicate neighbors. The columns following each neighborhood column are samples from that neighborhood.}
\label{fig:cluster}
\end{figure*}

To explore the sensitivity of our neighborhood based models, we inspect the behavior of our model using far fewer neighborhoods.
Another benefit of using fewer neighborhoods is that we can easily marginalize out the neighborhood and get marginal likelihoods. In this experiment, we cluster the training data into a small number of neighborhoods. For CelebA, we use kmeans to cluster the features extracted from $\beta$-VAE into 1000 neighborhoods. Five prototypes closest to the cluster center are chosen to condition the likelihood in the corresponding neighborhood. For MNIST, SVHN and CIFAR-10, we cluster the PCA features into 100, 500 and 1000 neighborhoods respectively. We use 5 prototypes for MNIST and SVHN, and 10 prototypes for CIFAR-10.

We conduct experiments using \textbf{NCT} here. Since the number of neighborhoods are fewer in this setting, we can exactly marginalize out the neighborhood. Training is still based on the neighbor conditioned objective $\log p_\theta(x_i \mid \textbf{N}(x_i))$. We report both neighbor conditioned likelihood and marginal likelihood for the test set in Tabel.~\ref{tab:cluster_likel}. We observe slightly higher bit per dimension compared to results from Table.~\ref{tab:results} because of the higher diversity presented in each cluster, but the samples still present the attributes specified by the prototypes, as shown in Fig.~\ref{fig:cluster}. Moreover, the quantitative results suggest a significant improvement over vanilla RealNVP, as shown in Tabel.~\ref{tab:cluster_fid}. Note further that the quantitative results are competitive with the ones from nearest neighbors reported in Table.~\ref{tab:fid} and ~\ref{tab:prd}.

\begin{table}[h!]
\caption{Comparison between neighborhood conditioned likelihood and the marginal likelihood for \textbf{NCT} model. Results are presented in bpd.}
\label{tab:cluster_likel}
\centering
\begin{tabular}{lcccr}
\toprule
     Dataset & $p_{\theta}(x \mid \textbf{N}(x))$ &  $\frac{1}{N}\sum_{i=1}^{N}p_\theta(x\mid \textbf{N}(x))$\\
\midrule
     MNIST & 0.444 & 0.465\\
     SVHN & 2.217 & 2.286\\
     CIFAR-10 & 3.587 & 3.675\\
     CelebA & 3.073 & 3.075\\
\bottomrule
\end{tabular}
\end{table}

\begin{table}[h!]
\caption{FID and PRD scores for \textbf{NCT} model using cluster based neighborhoods.}
\label{tab:cluster_fid}
\centering
\begin{tabular}{lc|cr}
\toprule
     Dataset &  FID & PRD\\
\midrule
     MNIST & 10.8 & (0.980, 0.966)\\
     SVHN & 44.9 & (0.848, 0.951)\\
     CIFAR-10 & 84.6 & (0.476, 0.658)\\
     CelebA & 29.6 & (0.879, 0.828)\\
\bottomrule
\end{tabular}
\end{table}

\end{document}